\documentclass[runningheads]{llncs}
\usepackage[T1]{fontenc}
\usepackage{graphicx}
\usepackage{booktabs}
\usepackage{array}
\usepackage{subfigure}
\usepackage{caption}
\usepackage{float}
\usepackage[misc]{ifsym}
\usepackage[colorlinks]{hyperref}
\usepackage{multirow}
\usepackage{url}
\urldef{\mailsa}\path|chunfeng.lian@xjtu.edu.cn|
\urldef{\mailsb}\path|fan.wang@xjtu.edu.cn|
\urldef{\mailsc}\path|wzy218@xjtu.edu.cn|

\usepackage{graphicx}
\usepackage{amssymb}
\usepackage{amsmath}
\usepackage{mathrsfs}
\usepackage{bbding}
\usepackage{pifont}
\usepackage{threeparttable}
\usepackage{color}

\begin{document}
\title{Forensic Histopathological Recognition via a Context-Aware MIL Network Powered by Self-Supervised Contrastive Learning}
\titlerunning{FPath}

%\author{Anonymous Authors}

\author{Chen Shen\inst{1} \and Jun Zhang \inst{4}  \and Xinggong Liang\inst{1} \and Zeyi Hao\inst{1} \and Kehan Li\inst{3} \and  Fan Wang\inst{3}\textsuperscript{(\Letter)} \and Zhenyuan Wang \inst{1}\textsuperscript{(\Letter)} \and Chunfeng Lian\inst{2}\textsuperscript{(\Letter)}}
%index{Shen Chen \and Zhang Jun \and Liang XingGong \and Hao Zeyi \and Li Kehan \and Wang Fan \and Wang Zhenyuan \and Lian Chunfeng}
\authorrunning{C. Shen et al.}

%\institute{Submission ID: 944}

\institute{Key Laboratory of National Ministry of Health for Forensic Sciences, School of Medicine \& Forensics, Health Science Center, Xi'an Jiaotong University, Xi'an, Shaanxi 710049, China\\
\mailsc  \and
	School of Mathematics and Statistics, Xi'an Jiaotong University, Xi'an, Shaanxi 710149, China\\
\mailsa\and
	Key Laboratory of Biomedical Information Engineering of Ministry of Education, School of Life Science and Technology, Xi'an Jiaotong University, Xi'an, Shaanxi 710049, China \\
\mailsb\and
Tencent AI Lab, Shenzhen, China}

\maketitle              % typeset the header of the contribution
\begin{abstract}
Forensic pathology is critical in analyzing death manner and time from the microscopic aspect to assist in the establishment of reliable factual bases for criminal investigation.
In practice, even the manual differentiation between different postmortem organ tissues is challenging and relies on expertise, considering that changes like putrefaction and autolysis could significantly change typical histopathological appearance.
Developing AI-based computational pathology techniques to assist forensic pathologists is practically meaningful, which requires reliable discriminative representation learning to capture tissues' fine-grained postmortem patterns.
To this end, we propose a framework called FPath, in which a dedicated self-supervised contrastive learning strategy and a context-aware multiple-instance learning (MIL) block are designed to learn discriminative representations from postmortem histopathological images acquired at varying magnification scales.
Our self-supervised learning step leverages multiple complementary contrastive losses and regularization terms to train a double-tier backbone for fine-grained and informative patch/instance embedding.
Thereafter, the context-aware MIL adaptively distills from the local instances a holistic bag/image-level representation for the recognition task.
On a large-scale database of $19,607$ experimental rat postmortem images and $3,378$ real-world human decedent images, our FPath led to state-of-the-art accuracy and promising cross-domain generalization in recognizing seven different postmortem tissues.
The source code will be released on  \href{https://github.com/ladderlab-xjtu/forensic_pathology}{https://github.com/ladderlab-xjtu/forensic\_pathology}.
\end{abstract}

\vspace{-15pt}
\section{Introduction} \label{sec:intro}
%\vspace{-5pt}
Computational pathology powered by artificial intelligence (AI) shows promising applications in various clinical studies~\cite{RN115,RN146}, significantly easing the workload and promoting the development of clinical pathology.
Inspired by such exciting progress, let's think step by step, so why not leverage advanced AI techniques to boost the research and applications in another important discipline, i.e., forensic pathology?
Forensic pathology focuses on investigating the cause, manner, and time of (non-natural) deaths based on histopathological examinations of postmortem organ tissues~\cite{RN154}.
As an indispensable part of the medicolegal autopsy, it provides critical evidence from the microscopic aspect to confirm, perfect, or refute macroscopic findings, establishing a reliable factual basis for future inferences~\cite{RN158}.
Histopathological analysis in forensic pathology is challenging and time-consuming, since postmortem changes (e.g., putrefaction and autolysis) severely destroy tissues' typical image appearance, even making the manual differentiation between the tissues of different organs very difficult.
%Therefore, developing accurate and efficient computational pathology methods to assist forensic pathologists is urgently needed.

Although diverse deep-learning approaches have been proposed in clinical studies to process and analyze histopathological images~\cite{RN115,RN146}, no similar work has yet in the forensic pathology community.
The main reason could be three-fold.
\textbf{1)} Forensic and clinical pathology have distinct purposes. The former case analyzes the tissue images from multiple organs concurrently. In contrast, clinical diagnosis/prognosis usually focuses on one tissue type in one task~\cite{RN145}.
\textbf{2)} Due to postmortem changes, histopathological images in forensic pathology have atypical appearances and more complex distributions than in clinical pathology, bringing additional challenges to deep representation learning~\cite{wanggongji2022,RN204}.
\textbf{3)} Data in forensic pathology are more difficult to obtain and have relatively lower quality.
Therefore, to deploy a reliable computational pathology system for forensic investigation, fine-grained discriminative representation learning from complex postmortem histopathological images is a very precondition.

\begin{figure}[t]
\setlength{\abovecaptionskip}{-1pt}
\setlength{\belowcaptionskip}{-15pt}
  \centering
   \includegraphics[width=1.0\textwidth]{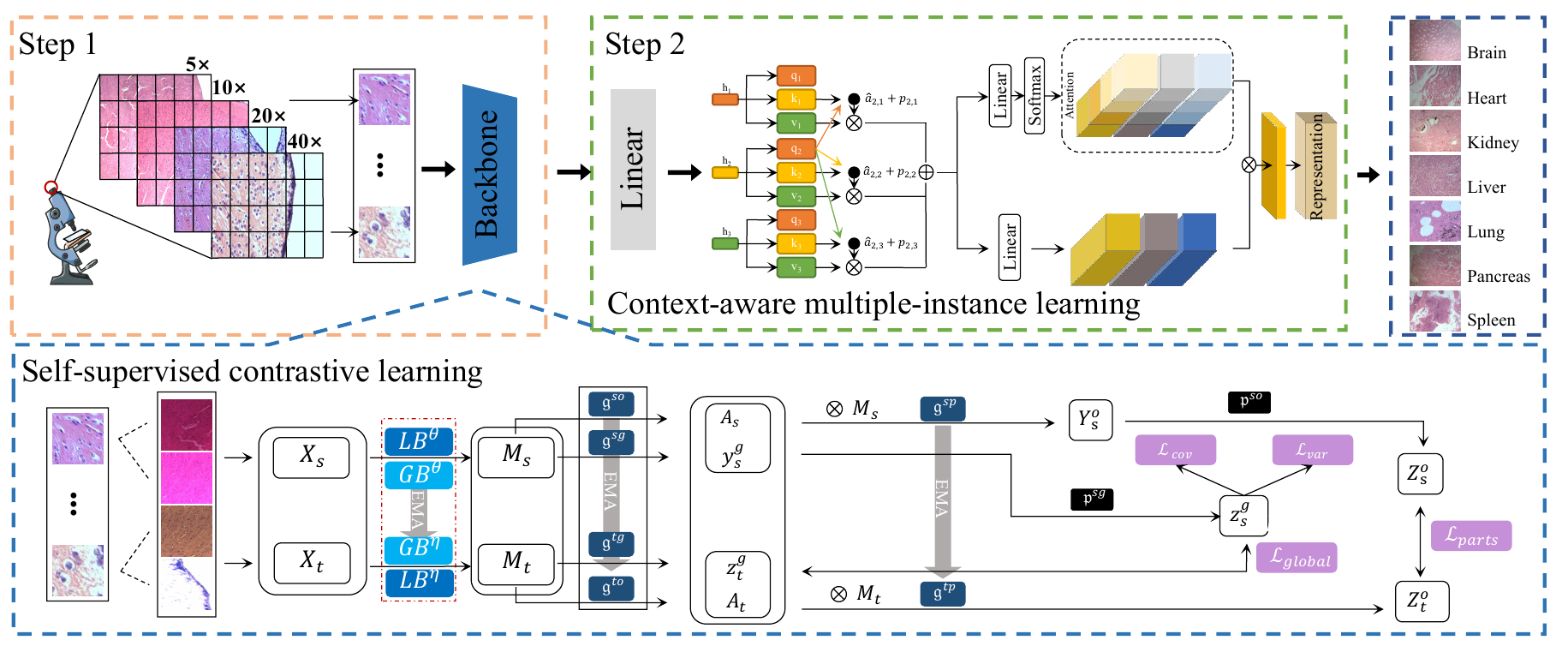}
  \caption{Our FPath that consists of a self-supervised double-tier backbone (Step 1) and a context-aware MIL branch for postmortem recognition (Step 2). }
\label{fig:overview}
\end{figure}

In this paper, we introduce a deep computational pathology framework (\emph{dubbed as} \textbf{FPath}) for forensic histopathological analysis.
As shown in Fig.~\ref{fig:overview}, FPath leverages the idea of self-supervised contrastive learning and multiple instance learning (MIL) to learn discriminative histopathological representations.
Specifically, we propose a self-supervised contrastive learning strategy to learn a double-tier backbone network for fine-grained feature embedding of local image patches (i.e., instances in MIL).
After that, a context-aware MIL block is designed, which adopts a self-attention mechanism to refine instance-level representations by aggregating contextual information, and then applies an adaptive-pooling operation to produce a holistic image-level representation for prediction.
Our FPath performs efficient predictions without the need for tedious pre-processing (e.g., foreground extraction/segmentation).
To the best of our knowledge, this paper is the first attempt that shows promising applications of advanced AI techniques (e.g., self-supervised contrastive learning) to forensic pathology.
%unlike the majority of existing studies in clinical pathology~\cite{RN115,2021arXiv210600908S} that typically use a pre-trained convolutional neural network (CNN),

The main technical contributions of our work are:
\vspace{-10pt}
\begin{itemize}
	\item [\textbf{1)}] We design a double-tier backbone and a dedicated self-supervised learning strategy to capture discriminative instance-level histopathological patterns of postmortem organ tissues.
The double-tier backbone combines CNN and transformer for local and non-local information fusion.
To effectively train such a backbone to handle images acquired with varying microscopic magnifications, the dedicated self-supervised learning strategy leverages multiple complementary contrastive losses and regularization terms to concurrently maximize global and spatially fine-grained similarities between different views of the same instances/patches in an informative representation space.
% multiple loss functions (instance loss, objective loss, variance loss, and covariance loss).

	\item  [\textbf{2)}] We design a context-aware MIL branch to produce the bag-level discriminative representations for accurate and efficient postmortem histopathological recognition.
Our MIL branch first refines instance embedding by leveraging a self-attention mechanism integrating positional embedding to model cross-patch associations for contextual information enhancement.
 Thereafter, an adaptive pooling operation is designed to learn deformable spatial attention to distill from contextually enhanced patch-level representations a holistic image-level representation for recognition.

	\item[\textbf{3)}] Our FPath was applied to recognize postmortem organ tissues, a fundamental task in forensic pathology.
To this end, we established a relatively large-scale multi-domain database consisting of an experimental rat postmortem dataset and a real-world human decedent dataset, each with $19,607$ and $3,378$ images acquired at a specific microscopic magnification (e.g., 5$\times$, 10$\times$, 20$\times$, and 40$\times$), respectively.
On such a multi-domain database, our FPath led to promising cross-domain generalization and state-of-the-art accuracy in recognizing seven different postmortem organs.
\end{itemize}

%%%%%%%%%%%%%%%%%%%%%%%%%%%%%%%%%%%%%%%%%
%%%%%%%%%%%%%%%%%%%%%%%%%%%%%%%%%%%%%%%%%

\vspace{-20pt}
\section{Method} \label{sec:method}
\vspace{-8pt}
The schematic diagram of our FPath is shown in Fig.~\ref{fig:overview}, which consists of two steps: \textbf{1)} Self-supervised contrastive learning of a double-tier backbone, and
\textbf{2)} Context-aware multiple instance learning for postmortem tissue recognition.

%The schematic diagram of our FPath is shown in Fig.~\ref{fig:overview}.
%It consists of two key components. \textbf{1)} Self-supervised contrastive learning of a double-tier backbone (i.e., Step 1 in Fig.~\ref{fig:overview}) for fine-grained instance embedding of image patches with varying magnification scales.
%\textbf{2)} Context-aware multiple instance learning (i.e., Steps 2 in Fig.~\ref{fig:overview}) to refine and aggregates instance representations to produce a holistic bag/image-level representation for postmortem tissue recognition.

\vspace{-12pt}
\subsection{Self-Supervised Contrastive Patch Embedding}

\vspace{-5pt}
\subsubsection{Double-Tier Backbone}
Given patches from a postmortem histopathological image acquired at a specific magnification (i.e., 5$\times$, 10$\times$, 20$\times$, or 40$\times$), we adopt a backbone with a local branch (LB) and a global branch (GB) for instance/patch feature embedding.
The LB is a ResNet50~\cite{7780459} consisting of $16$ successive bottlenecks, each with three convolutional layers with the kernel size of $1\times1$, $3\times3$, and $1\times1$, respectively.
The GB is a Swin Transformer~\cite{2021arXiv210314030L} that contains of a series of $12$ window-based multi-head self-attention modules.
Let an input patch be $\mathbf{X}\in\mathbb{R}^{H\times W\times3}$.
The corresponding feature embedding produced by the double-tier backbone will be $\mathbf{M}=\mathbf{M}_{\text{LB}}\oplus\mathbf{M}_{\text{GB}}$ ($\in \mathbb{R}^{h\times w\times C}$), where $\mathbf{M}_{\text{LB}}$ and $\mathbf{M}_{\text{GB}}$ denotes the representations from the LB and GB branch, respectively, and $\oplus$ stands for the channel-wise concatenation operation.

\vspace{-17pt}
\subsubsection{Self-Supervised Contrastive Learning Strategy}
We leverage the idea of self-supervised representation learning to establish the double-tier backbone.
Referring to MoCo~\cite{2019arXiv191105722H}, our self-supervised learning is constructed by a teacher branch and a student branch.
The student branch consists of six components, including a double-tier backbone (i.e., $\mathfrak{f}_{\theta}(\cdot)$), three projection layers (i.e., $\mathfrak{g}^{\text{sg}}(\cdot)$, $\mathfrak{g}^{\text{so}}(\cdot)$, $\mathfrak{g}^{\text{sp}}(\cdot)$), and two prediction layers (i.e., $\mathfrak{p}^{\text{sg}}(\cdot)$ and $\mathfrak{p}^{\text{so}}(\cdot)$).
The teacher branch contains four components, including a double-tier backbone $\mathfrak{f}_{\eta}(\cdot)$, and three projection layers (i.e., $\mathfrak{g}^{\text{tg}}(\cdot)$, $\mathfrak{g}^{\text{to}}(\cdot)$, and $\mathfrak{g}^{\text{tp}}(\cdot)$).
By feeding the two branches with different views of same patches, $\mathfrak{f}_{\theta}(\cdot)$ in the student branch (i.e., parameterized by $\theta$) is trained via back-propagation to update $\mathfrak{f}_{\eta}(\cdot)$ in the teacher branch (i.e., parameterized by $\eta$) in a momentum-based moving average fashion, such as $\eta \gets m\cdot\eta + (1-m)\cdot\theta$, where $m=0.99$ is the momentum parameter.

Another key issue that determines the quality of the embedding from such a self-supervised strategy is the formulation of respective contrastive loss functions and regularization terms.
Accordingly, we design a thorough contrastive learning strategy to capture fine-grained discriminative patterns of postmortem tissues under varying microscopic magnifications.
That is, let $\mathbf{X}_{\text{s}}$ and $\mathbf{X}_{\text{t}}$ be two different views of an image patch $\mathbf{X}$ generated by a random data augmentation process.
Our contrastive learning strategy concurrently encourages the \emph{global similarity} and \emph{spatially fine-grained similarity} between the corresponding feature embedding $\mathbf{M}_{\text{s}}=\mathfrak{f}_{\theta}(\mathbf{X}_{\text{s}})$ and $\mathbf{M}_{\text{t}}=\mathfrak{f}_{\eta}(\mathbf{X}_{\text{t}})$ ($\in \mathbb{R}^{h\times w\times C}$).
Also, \emph{two regularization terms} are applied as auxiliary guidance to \emph{protect the informativeness and avoid collapses} of the embedding learned by the backbone.
% that concurrently minimizes multiple complementary loss functions and regularization terms

Specifically, the global similarity between $\mathbf{M}_{\text{s}}$ and $\mathbf{M}_{\text{t}}$ is encouraged by minimizing a general cosine contrastive loss, such as
{\setlength\abovedisplayskip{0.1cm}
\setlength\belowdisplayskip{0.1cm}
\begin{equation}\label{equ:global}
	\mathcal{L}_{\text{global}}=2-2\cdot\frac{<\mathbf{z}^{\text{g}}_{\text{s}}, \mathbf{z}^{\text{g}}_{\text{t}}>}{{||\mathbf{z}^{\text{g}}_{\text{s}}||}_2\cdot{||\mathbf{z}^{\text{g}}_{\text{t}}||}_2},
\end{equation}
}
where $\mathbf{z}^{\text{g}}_{\text{s}} = \mathfrak{p}^{\text{sg}}(\mathfrak{g}^{\text{sg}}(\text{GAP}(\mathbf{M}_{\text{s}})))$ and $\mathbf{z}^{\text{g}}_{\text{t}} = \mathfrak{g}^{\text{tg}}(\text{GAP}(\mathbf{M}_{\text{t}}))$, with $\text{GAP}(\cdot)$
standing for the global average pooling that produces feature vectors.

In practice, forensic pathologists typically infer postmortem tissue type by evaluating the cellular compositions in multiple local regions.
%, suggesting that fine-grained histopathological patterns could be informative in differentiation between different tissues.
Accordingly, inspired by cross-view learning~\cite{huang2022learning}, we design a spatially fine-grained contrastive loss to explicitly encourage multi-parts similarity between $\mathbf{M}_{\text{s}}$ and $\mathbf{M}_{\text{t}}$.
Assume $\mathbf{M}'_{\text{s}}$ and $\mathbf{M}'_{\text{t}}$ are two $(h\cdot w)\times C$ tensors flattened from $\mathbf{M}_{\text{s}}$ and $\mathbf{M}_{\text{t}}$ across the spatial dimension, respectively. They are further processed by $\mathfrak{g}^{\text{so}}(\cdot)$ and $\mathfrak{g}^{\text{to}}(\cdot)$ (followed by softmax normalization), respectively, to produce two $(h\cdot w)\times K$ attention matrices, i.e., $\mathbf{A}_{\text{s}}=\mathfrak{g}^{\text{so}}(\mathbf{M}'_{\text{s}})$ and $\mathbf{A}_{\text{t}}=\mathfrak{g}^{\text{to}}(\mathbf{M}'_{\text{t}})$, where $K$ denotes the predefined number of parts.
Thereafter, we aggregate the backbone representations in terms of the attention matrices to deduce multi-parts representations, i.e., $\mathbf{Z}^{\text{o}}_{\text{s}}=\mathfrak{p}^{\text{so}}(\mathfrak{g}^{\text{sp}}(\mathbf{A}_{\text{s}}^T\otimes \mathbf{M}'_{\text{s}}))$ and $\mathbf{Z}^{\text{o}}_{\text{t}}=\mathfrak{g}^{\text{tp}}(\mathbf{A}_{\text{t}}^T\otimes \mathbf{M}'_{\text{t}})$, where $\otimes$ denotes tensor multiplication.
Finally, the spatially fine-grained contrastive loss is quantified as
{\setlength\abovedisplayskip{0cm}
\setlength\belowdisplayskip{0cm}
\begin{equation}\label{equ:parts}
\mathcal{L}_{\text{parts}} = \sum_{k=1}^K \left(2-2\cdot\frac{<\mathbf{Z}^{\text{o}}_{\text{s}}[k,:], \mathbf{Z}^{\text{o}}_{\text{t}}[k,:]>}{{||\mathbf{Z}^{\text{o}}_{\text{s}}[k,:]||}_2\cdot{||\mathbf{Z}^{\text{o}}_{\text{t}}[k,:]||}_2} \right)
\end{equation}
}
where $\mathbf{Z}^{\text{o}}[k,:]$ denotes the $k$th part representation in $\mathbf{Z}^{\text{o}}\in \mathbb{R}^{K\times D}$.

Besides, two additional regularization terms are further included to stabilize contrastive representation learning. Following~\cite{bardes2021vicreg}, we penalize small changes between the global representations of different image patches across each feature dimension.
Also, we encourage the global representations to be diverse/orthogonal across different feature dimensions. Let $\mathcal{Z}_{\text{s}}^{\text{g}}$ be a set of feature representations for an input mini-batch of patches in the student branch, and $\widetilde{\mathcal{Z}_{\text{s}}^{\text{g}}}$ and $\overline{\mathcal{Z}_{\text{s}}^{\text{g}}}$ denote their channel-wise variation and mean.
The regularization terms are defined as
{\setlength\abovedisplayskip{0cm}
\setlength\belowdisplayskip{0cm}
\begin{equation}\label{equ:var}
	\mathcal{L}_{\text{var}}=\frac{1}{D}\sum_{d=1}^D \max\left(0, 1-\sqrt{\widetilde{\mathcal{Z}_{\text{s}}^{\text{g}}}[d]+\epsilon}\right)
\end{equation}}
{\setlength\abovedisplayskip{0cm}
\setlength\belowdisplayskip{0cm}
\begin{equation}\label{equ:cov}
	\mathcal{L}_{\text{cov}}=\frac{1}{D^2-D}\sum_{i\neq j}\left(\left\{(\mathbf{z}^{\text{g}}_{\text{s}}-\overline{\mathcal{Z}_{\text{s}}^{\text{g}}})^T(\mathbf{z}^{\text{g}}_{\text{s}}-\overline{\mathcal{Z}_{\text{s}}^{\text{g}}})\right\}[i,j]\right)^2
\end{equation}}
where $\epsilon$ is a small scalar to stabilize numerical computation, $\widetilde{\mathcal{Z}_{\text{s}}^{\text{g}}}[d]$ denotes the $d$th dimension of $\widetilde{\mathcal{Z}_{\text{s}}^{\text{g}}}$, and $\{(\mathbf{z}^{\text{g}}_{\text{s}}-\overline{\mathcal{Z}_{\text{s}}^{\text{g}}})^T(\mathbf{z}^{\text{g}}_{\text{s}}-\overline{\mathcal{Z}_{\text{s}}^{\text{g}}})\}[i,j]$ is the $[i,j]$th element in such a covariance matrix.
According to~\cite{bardes2021vicreg}, Eqs.~(\ref{equ:var}) and~(\ref{equ:cov}) jointly encourage the diversity across patches and feature dimensions, thus protecting the informativeness and avoid collapse of self-supervised contrastive learning.

Overall, we combine Eqs.~(\ref{equ:global}) to~(\ref{equ:cov}) as the final loss function to train the double-tier backbone, such as
$ \mathcal{L}_{\text{all}}=\mathcal{L}_{\text{global}}+\mathcal{L}_{\text{parts}}+\gamma\mathcal{L}_{\text{var}}+\lambda\mathcal{L}_{\text{cov}}
$,
where $\gamma$ and $\lambda$ are two tuning parameters balancing different terms.

\vspace{-12pt}
\subsection{Context-Aware MIL}
%\vspace{-5pt}
Given the patch/instance-level representations of a histopathological image from the double-tier backbone, we further design a context-aware MIL framework to aggregate their information for postmortem tissue recognition.
Given patch embeddings of a Microscope image, our context-aware MIL part contains two main steps, i.e., a multi-head self-attention to refine each patch’s feature and an adaptive pooling step to distill all patches’ information.

In detail, we first adopt a multi-head self-attention (MSA) mechanism~\cite{shao2021transmil} integrating relative positional embedding to explicitly model cross-patch associations for contextual enhancement of the instance representations from the backbone.
Let $\mathcal{Z}=\{\mathbf{z}_i\}_{i=1}^I$ be a set of the contextually enhanced instance embedding from an image.
Thereafter, inspired by Deformable DETR~\cite{2020arXiv201004159Z}, we further design an adaptive pooling operation, which is simple but effective to distill from $\mathcal{Z}$ a bag-level holistic representation for the classification purpose.
Specifically, the bag-level holistic representation determined by the adaptive pooling is
{\setlength\abovedisplayskip{0cm}
\setlength\belowdisplayskip{0cm}
 \begin{equation}\label{equ:z_bag} \mathbf{z}_{\text{bag}}=\frac{1}{I}\sum_{i=1}^I(softmax(\mathfrak{h}_{\omega_1}(\mathbf{z}_i))\circ\mathfrak{h}_{\omega_2}(\mathbf{z}_i)) ,
\end{equation}}
where $\mathfrak{h}_{\omega_1}(\cdot)$ and $\mathfrak{h}_{\omega_2}(\cdot)$ are two linear projections with the same number of output units, symbol $\circ$ denotes the Hadamard product between two tensors, and $softmax(\cdot)$ is performed across different instances to filter out uninformative patches and preserve discriminative patches in quantifying $\mathbf{z}_{\text{bag}}$ for classification.
%Finally, the bag-level representation $\mathbf{z}_{\text{bag}}$ is fed into a linear classifier to determine the tissue type of the input postmortem histopathological image.

\vspace{-12pt}
\section{Experiments}

\vspace{-8pt}
\subsection{Data \& Experimental Setup}

\vspace{-5pt}
\subsubsection{Rat postmortem histopathology dataset}
Ninety Sprague-Dawley adult male rats were executed by the spinal cord dislocation and placed in a constant temperature and humidity environment for 6-8 hours.
The animal experiments were approved by the Laboratory Animal Care Committee of the anonymous institution.
Seven organs, i.e., brain, heart, kidney, liver, lung, pancreas, and spleen, were removed and placed in the formalin solution.
Briefly, paraffin sections of these organ tissues were stained with the H\&E solution.
The H\&E-stained sections were then analyzed by three forensic pathologists, who used Lercai LAS EZ microscopes to record the areas according to their expertise.
Overall, five to ten images were recorded from a section at each magnification (i.e., $5\times$, $10\times$, $20\times$, and $40\times$).
Finally, we split the 90 rats as training, validation, and test sets of 60, 10, and 20 rats, respectively, each with $13,137$, $2,235$, and $4,325$ images.

\vspace{-15pt}
\subsubsection{Human forensic histopathology dataset}
The real forensic images were provided by the Forensic Judicial Expertise Center of the anonymous institution, after getting the informed consent of relatives.
All procedures followed the requirements of local laws and institutional guidelines, and were approved and supervised by the Ethics Committee.
A total of 32 decedents participated in this study.
Four to six images were recorded at each of three magnifications ($5\times$, $10\times$, and $20\times$) per H\&E stained section.
Similar to the rat dataset, the human dataset was selected from the same seven organs.
Finally, the training, validation and test sets contain  $1,691$ images, $628$, and $1059$ images, corresponding to 16, 6, and 10 different decedents, respectively.

\vspace{-15pt}
\subsubsection{Experimental details}
Notably, the double-tier backbone was self-supervised and learned on the rat training set for 100 epochs by setting the mini-batch size as 1024, with the parameters initialized by the ImageNet pre-trained models.
The training data were augmented by a histopathology-oriented strategy by combining different kinds of staining jitters, random affine transformation, Gaussian blurring, resizing, etc.
The image(patch) dimension in our implementation was 224*224.
%Following~\cite{2020arXiv200607733G}, LARS\cite{2017arXiv170803888Y} was adopted as the optimizer with an initial learning rate of $0.8$, which was updated by cosine decay schedule~\cite{2016arXiv160803983L}.
The tuning parameters $\gamma$  and $\lambda$ in $\mathcal{L}_{\text{all}}$ were set as 5 and 0.005, respectively.
Thereafter, the MIL blocks on two different datasets were both trained by minimizing the cross-entropy loss for 20 epochs with the mini-batch size setting as 32.
The experiments were conducted on three PCs with twenty NVIDIA GEFORCE RTX 3090 GPUs.
% via momentum-based SGD~\cite{2016arXiv160904747R}, with an initial learning rate of 0.04 and momentum of 0.9.

\vspace{-10pt}
\subsection{Results of Self-Supervised Contrastive Learning}
Our self-supervised double-tier backbone was compared with other state-of-the-art self-supervised learning approaches, including \textbf{balow twins}~\cite{zbontar2021barlow}, \textbf{swin transformer (SSL)}~\cite{xie2021moby}, \textbf{TransPath}~\cite{wang2021transpath}, \textbf{CTransPath}~\cite{wang2022},\textbf{RetCCL}~\cite{WANG2023retccl}  and \textbf{MOCOV3}~\cite{2021arXiv210402057C}.
To evaluate the discriminative power of  these competing methods, we adopted GAP to aggregate their instance representations from a whole image to train simple linear classifiers for the recognition of seven different organ tissues on both the rat and human datasets, with the test performance quantified in terms of four general classification metrics (i.e., \textbf{ACC}, \textbf{F1 score}, \textbf{MCC(Matthews Correlation Coefficient)}, and \textbf{Precision}). The corresponding results are summarized in Table~\ref{tab:ssl-compare}, from which we can have two observations. \emph{First}, our self-supervised double-tier backbone consistently outperformed all other competing methods in terms of all metrics on two datasets.
\emph{Second}, our method led to better generalization, as the backbone trained on the rat dataset shows promising performance on the challenging real-world human dataset (e.g., resulting in an ACC higher than $90\%$).
These results suggest the effectiveness of our self-supervised learning strategy.

\begin{table}[t]
\centering
\setlength\belowdisplayskip{-15pt}
	\caption{Linear classification results obtained by different self-supervised learning approaches on the rat and human testing sets, respectively.}
	\label{tab:ssl-compare}
	\resizebox{\textwidth}{!}{
			\begin{tabular}{c|cccc|cccc}
				\hline
\multirow{2}{*}{\textbf{Competing methods}}			& \multicolumn{4}{c|}{\textbf{Rat dataset}}                                        & \multicolumn{4}{c}{\textbf{Human dataset}}                                        \\
				& ACC             & F1              & MCC             & Precision             & ACC             & F1             & MCC             & Precision             \\ \hline
				balow twins~\cite{zbontar2021barlow}           & 0.9232          & 0.9123          & 0.9076          & 0.9070          & 0.7306          & 0.7311          & 0.6854          & 0.7345          \\
				swin transformer(SSL)~\cite{xie2021moby} & 0.9450          & 0.9369          & 0.9299          & 0.9330          & 0.8079          & 0.8088          & 0.7758          & 0.8125          \\
				Transpath~\cite{wang2021transpath}             & 0.7351          & 0.7397          & 0.6958          & 0.7481          & 0.5838          & 0.5657          & 0.5264          & 0.6050          \\
				CTransPath~\cite{wang2022}            & 0.9635          & 0.9610          & 0.9535          & 0.9596          & 0.8794          & 0.8799          & 0.8591          & 0.8842          \\
				RetCCL~\cite{WANG2023retccl}            & 0.9794          & 0.9801         & 0.9768          & 0.9810          & 0.7796          & 0.7789          & 0.7448          & 0.7961          \\
				MOCOV3~\cite{2021arXiv210402057C}                & 0.9732          & 0.9738          & 0.9681          & 0.9745          & 0.8103          & 0.8124          & 0.7790          & 0.8187          \\
				Ours                   & \textbf{0.9831} & \textbf{0.9831} & \textbf{0.9796} & \textbf{0.9831} & \textbf{0.9049} & \textbf{0.9044} & \textbf{0.8886} & \textbf{0.9056} \\ \hline
			\end{tabular}
	}
\vspace{-5pt}
\centering
	\caption{Ablation studies to evaluate the contributions of different self-supervised contrastive losses and regularization terms.}
	\label{tab:ssl-ablation}
	\resizebox{\textwidth}{!}{
		\begin{tabular}{cccc|cccc|cccc}
			\hline
	\multicolumn{4}{c|}{\textbf{Loss functions}}     & \multicolumn{4}{c|}{\textbf{Rat dataset}}                                          & \multicolumn{4}{c}{\textbf{Human dataset}}                                        \\
			 $\mathcal{L}_{\text{global}}$ & $\mathcal{L}_{\text{parts}}$ & $\mathcal{L}_{\text{var}}$ & $\mathcal{L}_{\text{cov}}$ & ACC             & F1             & MCC             & Precision             & ACC             & F1              & MCC             & Precision             \\ \hline
			  $\checkmark$    &     &     &     & 0.9732          & 0.9713          & 0.9660          & 0.9689          & 0.8918          & 0.8913          & 0.8734          & 0.8935          \\
			 $\checkmark$    & $\checkmark$    &     &     & 0.9817          & 0.9819          & 0.9778          & 0.9822          & 0.8953          & 0.8956          & 0.8779          & 0.8981          \\
			 $\checkmark$   &     & $\checkmark$    & $\checkmark$    & 0.9793          & 0.9799          & 0.9757          & 0.9806          & 0.8978          & 0.8976          & 0.8802          & 0.8983          \\
			$\checkmark$    & $\checkmark$    & $\checkmark$    & $\checkmark$    & \textbf{0.9831}          & \textbf{0.9831}          & \textbf{0.9796  }        & \textbf{0.9831} & \textbf{0.9049} & \textbf{0.9044} & \textbf{0.8886} & \textbf{0.9056} \\ \hline
		\end{tabular}
}
\end{table}	

For a more detailed evaluation, we further conducted a series of ablation studies to evaluate the contributions of the contrastive losses (i.e., $\mathcal{L}_{\text{global}}$ and $\mathcal{L}_{\text{parts}}$) and regularization strategy (i.e., $\mathcal{L}_{\text{var}}+\mathcal{L}_{\text{cov}}$).
The corresponding results are summarized in Table~\ref{tab:ssl-ablation}, from which we can see that, given the baseline of $\mathcal{L}_{\text{global}}$, both the inclusion of the spatially fine-grained contrastive loss (i.e., $\mathcal{L}_{\text{parts}}$) and informativeness regularization (i.e., $\mathcal{L}_{\text{var}}$ and $\mathcal{L}_{\text{cov}}$) led to respective performance gains.
%For the results obtained by combinations of these losses and regularizations, please refer to the \emph{Supplementary Materials}.
These results further justify our self-supervised design.

\vspace{-10pt}
\subsection{Results of Multiple-Instance Learning }
Based upon the double-tier backbone learned on the rat training set, we compared our context-aware MIL with other MIL methods, including the gated attention-based approach (i.e., \textbf{AB-MIL}~\cite{2018arXiv180204712I},	\textbf{DSMIL}~\cite{li2021dual}, \textbf{Transmil}~\cite{shao2021transmil} and \textbf{MSA}~\cite{li2021dt,chen2022scaling}) approaches with/without different positional embedding strategies, i.e.,  relative position embedding (\textbf{MSA-RP}~\cite{2021arXiv210314030L}), learnable position embedding (\textbf{MSA-LP}~\cite{dosovitskiy2020image}), and 2D sine-cosine position embedding (\textbf{MSA-SP}~\cite{he2022masked}).
Notably, our approach used MSA-RP as the baseline, based on which an adaptive pooling operation is designed to produce the final bag-level representation.
To check the efficacy of \textbf{adaptive pool}, we further conducted a corresponding set of ablation studies by replacing it with other operations, including \textbf{max pool}, and \textbf{soft pool}~\cite{stergiou2021refining}.
These comparison and ablations results are shown in Table~\ref{tab:mil}, from which we can observe that our method led to the best results on both datasets, with relatively more significant improvements on the challenging human dataset.
Also, compared with other pooling operations, the adaptive pool design brought consistent performance gains.
These results suggest the efficacy of our context-aware MIL for postmortem tissue recognition.

\begin{table}[t]
\centering
	\caption{Multiple-instance learning results obtained by the competing methods and our Context-Aware MIL with different pooling strategies.}
	\label{tab:mil}
	\resizebox{\textwidth}{!}{
	\begin{tabular}{l|cccc|cccc}
		\hline
\multirow{2}{*}{\textbf{Competing methods}}			& \multicolumn{4}{c|}{\textbf{Rat dataset}}                                          & \multicolumn{4}{c}{\textbf{Human dataset}}                                        \\
		& ACC            & F1             & MCC             & Precision       & ACC             & F1             & MCC             & Precision       \\ \hline
		AB-MIL\cite{2018arXiv180204712I}                                                                            & 0.9815          & 0.9828          & 0.9793          & 0.9844          & 0.9011          & 0.9005          & 0.8838          & 0.9050          \\
	DSMIL~\cite{li2021dual}      & 0.9951          & 0.9948          & 0.9937          & 0.9945          & 0.9176          & 0.9166          & 0.9030          & 0.9170          \\
		Transmil~\cite{shao2021transmil}      & 0.9899          & 0.9888          & 0.9875          & 0.9878          & 0.8824          & 0.8813          & 0.8622          & 0.8821          \\		
	MSA~\cite{li2021dt,chen2022scaling}          & 0.9875          & 0.9883          & 0.9861          & 0.9892          & 0.9082          & 0.9082          & 0.8921          & 0.9100          \\
	MSA-LP~\cite{dosovitskiy2020image}      & 0.9879          & 0.9875          & 0.9853          & 0.9873          & 0.9097          & 0.9087          & 0.8945          & 0.9109          \\
		MSA-SP~\cite{he2022masked}  & 0.9851          & 0.9839          & 0.981           & 0.9832          & 0.8915          & 0.8905          & 0.8748          & 0.8948          \\
		MSA-RP~\cite{2021arXiv210314030L}      & 0.9915          & 0.9915          & 0.9896          & 0.9916          & 0.9218          & 0.9213          & 0.9085          & 0.9218          \\\hline
		Ours + Max pool                                                                       & 0.9910          & 0.9909          & 0.9888          & 0.9909          & 0.9144          & 0.9147          & 0.9001          & 0.9191          \\
		Ours + Soft pool~\cite{stergiou2021refining}                                                                      & 0.9935          & 0.9929          & 0.9915          & 0.9924          & 0.9047          & 0.9023          & 0.8883          & 0.9056          \\
		Ours + Adaptive pool                                                                  & \textbf{0.9956} & \textbf{0.9952} & \textbf{0.9943} & \textbf{0.9949} & \textbf{0.9229} & \textbf{0.9218} & \textbf{0.9093} & \textbf{0.9263} \\ \hline
	\end{tabular}
}
\end{table}
\begin{figure}[t]
\setlength{\abovecaptionskip}{-1pt}
\setlength{\belowcaptionskip}{-18pt}
	\centering
	\includegraphics[width=1.0\textwidth]{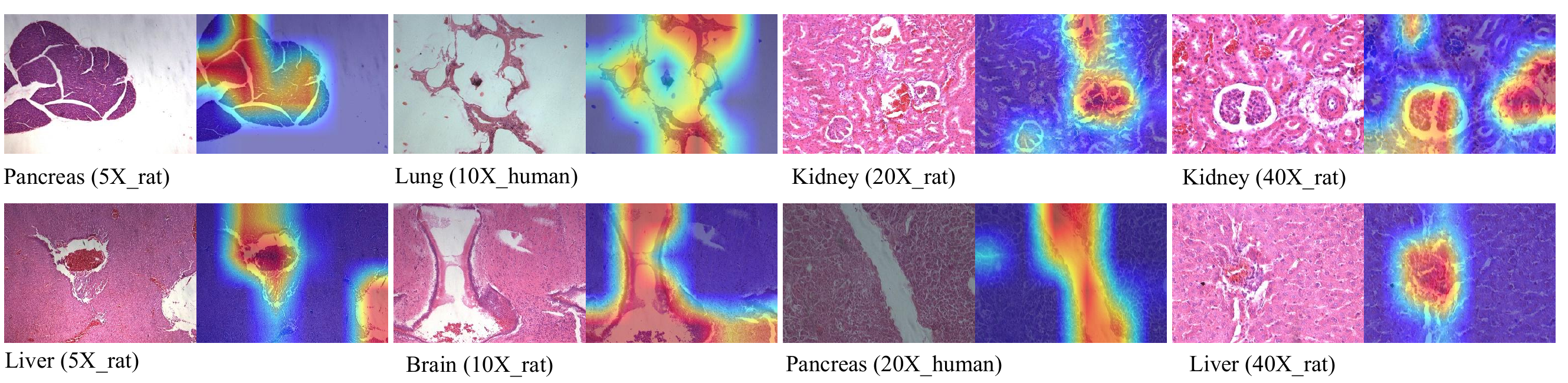}
	\caption{ Explainability analysis based on LayerCAM~\cite{jiang2021layercam} for representative postmortem tissue images acquired at different microscopic scales.}
	\label{fig2}
\end{figure}

In addition, we conducted LayerCAM-based analysis~\cite{jiang2021layercam} to check the explainability and reliability of our postmortem histopathological recognition results.
From the representative examples shown in Fig.~\ref{fig2}, we can have an interesting observation that our method tends to focus on tissue-specific postmortem patterns at different microscopic scales.
For example, the spatial attention maps reliably highlighted the meningeal structures of the brain tissue, the glomeruli in the kidney cortex, and the central vein area between the liver lobules.
On the other hand, based on the pancreas example, we can see that our network can sensitively localize the pancreas glandular structure while filtering out the uninformative background in an end-to-end fashion, without the need for any pre-processing to segment first the foreground.
These observations support our assumption that the proposed method is reliable and efficient in learning discriminative histopathological representations of postmortem organ tissues.

\vspace{-12pt}
\section{Conclusion}
\vspace{-8pt}
In this study, we have proposed a context-aware MIL framework powered by self-supervised contrastive learning to learn fine-grained discriminative representations for postmortem histopathological recognition.
 The dedicated self-supervised learning strategy concurrently maximizes multiple contrastive losses and regularization terms to deduce informative and discriminative instance embedding.
Thereafter, the context-aware MIL framework adopts MSA followed by an adaptive pooling operation to distill from all instances a holistic bag/image-level representation.
 The experimental results on a relatively large-scale database suggest the state-of-the-art postmortem recognition performance of our method.

\bibliographystyle{splncs04}
\bibliography{paper944}	
\end{document}